\documentclass[runningheads]{llncs}
\usepackage[T1]{fontenc}
\usepackage{graphicx}
\usepackage[numbers]{natbib}
\usepackage{amssymb}
\usepackage{booktabs}
\usepackage{graphicx}

\usepackage[nolist]{acronym}
\usepackage{color, colortbl}
\usepackage{paralist}
\usepackage{hyperref}
\usepackage{xurl}

\usepackage[skins]{tcolorbox}
\usepackage{xcolor}

\tcbset{
  my box/main style/.style={},
  my box/title style/.style={},
  my box/main/.style={/tcb/my box/main style/.style={#1}},
  my box/title/.style={/tcb/my box/title style/.style={#1}},
  my box/append main/.style={/tcb/my box/main style/.append style={#1}},
  my box/append title/.style={/tcb/my box/title style/.append style={#1}},
  my box/.style={
    my box/.cd, #1,
    /tcb/.cd,
    enhanced,
    my box/main style,
    attach boxed title to top left={xshift=0.2cm, yshift=-0.2cm},
    boxed title style={
      top=1pt,
      bottom=1pt,
      my box/title style,
    },
  },
}

\newtcolorbox{finding}[2][]{
  my box={
    main={
        top=6pt,
        bottom=3pt,
        colframe=black, 
        colback=white
    },
    title={
        colframe=black, 
        colback=gray
    },
  },
  title=#2,
  #1,
}

\newtcolorbox{principle}[2][]{
  my box={
    main={
        top=6pt,
        bottom=3pt,
        colframe=black, 
        colback=white
    },
    title={
        colframe=black, 
        colback=gray
    },
  },
  title=#2,
  #1,
}

\def\BibTeX{{\rm B\kern-.05em{\sc i\kern-.025em b}\kern-.08em
    T\kern-.1667em\lower.7ex\hbox{E}\kern-.125emX}}

\newcommand{\quotes}[1]{``#1''}

\begin{document}


\title{Illocutionary Explanation Planning for Source-Faithful Explanations in Retrieval-Augmented Language Models}
\titlerunning{Chain-of-Illocution Prompting}

\author{Francesco Sovrano\inst{1,2}\orcidID{0000-0002-6285-1041} \and
Alberto Bacchelli\inst{2}\orcidID{0000-0003-0193-6823}}
%
%
\institute{Collegium Helveticum, ETH Zurich, Zurich, Switzerland\\
\email{fsovrano@ethz.ch} \and
University of Zurich, Zurich, Switzerland\\
\email{\{sovrano,bacchelli\}@ifi.uzh.ch}}

\maketitle

\begin{abstract}
Natural language explanations produced by large language models (LLMs) are often persuasive, but not necessarily \emph{scrutable}: users cannot easily verify whether the claims in an explanation are supported by evidence. In XAI, this motivates a focus on \emph{faithfulness} and \emph{traceability}, i.e., the extent to which an explanation's claims can be grounded in, and traced back to, an explicit source.
We study these desiderata in retrieval-augmented generation (RAG) for programming education, where textbooks provide authoritative evidence. We benchmark six LLMs on 90 Stack Overflow questions grounded in three programming textbooks and quantify source faithfulness via source adherence metrics. We find that non \ac{RAG} models have median source adherence of 0\%, while baseline \ac{RAG} systems still exhibit low median adherence (22--40\%, depending on the model). 
Motivated by Achinstein's illocutionary theory of explanation, we introduce \emph{illocutionary macro-planning} as a descriptive design principle for source-faithful explanations and instantiate it with \emph{chain-of-illocution prompting} (\texttt{CoI}), which expands a query into implicit explanatory questions that drive retrieval.
Across models, \texttt{CoI} yields statistically significant gains (up to 63\%) in source adherence, although absolute adherence remains moderate and the gains are weak or non-significant for some models. A user study with 165 retained participants (220 recruited) indicates that these gains do not harm satisfaction, relevance, or perceived correctness.
\\\textbf{Data \& Code:} \href{https://github.com/Francesco-Sovrano/chain-of-illocution}{https://github.com/Francesco-Sovrano/chain-of-illocution}
\end{abstract}

\begin{acronym}
    \acro{AI}{Artificial Intelligence}
    \acro{GenAI}{Generative Artificial Intelligence}
    \acro{RAG}{Retrieval-Augmented Generation}
    \acro{LLM}{Large Language Model}
\end{acronym}

\section{Introduction} \label{sec:introduction}

\acp{LLM} are increasingly used as \emph{explainers} (from interactive assistants to tutoring systems) because they can produce fluent, context-sensitive natural language rationales \cite{kasneci2023chatgpt,tlili2023if}. Yet fluency is not enough: natural language explanations can be compelling even when they are unsupported by evidence. This matters whenever users must be able to \emph{audit} an explanation, rather than merely accept it.
In XAI, this gap is typically framed in terms of \emph{faithfulness} and \emph{traceability}: an explanation should reflect the reasoning or evidence that actually supports the output, and users should be able to trace claims back to an explicit source \cite{jacovi2020towards,sovrano2025legal}. In this work, we focus on traceability to \emph{external evidence} (rather than internal model states), and \textbf{we study how to design generation pipelines that keep natural language explanations anchored to their sources}.

Formal education provides a concrete testbed for these desiderata. Established curricula specify what learners should know and what counts as a correct solution \cite{DfE2024,glatthorn2004developing}, while textbooks and other peer-reviewed materials operationalize these standards in a stable, inspectable form \cite{hilton2016open}. This makes it possible to audit an explanation by checking whether its claims are warranted by an authoritative reference text. At the same time, the educational landscape is evolving rapidly, especially in programming and software engineering education: \ac{GenAI} tools like ChatGPT offer personalized, on-demand responses to queries \cite{liu2023chatgpt,da2024chatgpt}. Unlike textbooks, however, \ac{GenAI} outputs are unstable and prone to inaccurate or fabricated information (hallucinations). Recent research reports that up to 52\% of ChatGPT's responses on Stack Overflow contain inaccuracies, and 36\% include non-factual or fabricated content \cite{kabir2024stack}, contributing to platform policies restricting AI-generated answers \cite{stackoverflow2023policy}.

\ac{RAG} is a natural mitigation: it queries canonical textbooks for authoritative passages and conditions generation on the retrieved evidence. From an explainability perspective, however, retrieval is only a means to an end: the goal is a \emph{scrutable} explanation whose claims remain traceable to the evidence source. In retrieval-augmented settings, we operationalize this notion of explanation faithfulness as \emph{source adherence}: the extent to which the generated explanation is supported by the retrieved text. We therefore treat textbook adherence as an XAI-relevant property of natural language explanations, and first ask: \textbf{to what extent do widely used \ac{RAG} systems actually adhere to their reference sources?} (\textbf{RQ1})

To answer \textbf{RQ1}, we evaluated six RAG-enhanced \acp{LLM}: GPT-3.5-turbo, GPT-4o, Llama 3 (8B and 70B), Mistral, and Mixtral 8x7B. We tested these models on three programming education textbooks covering Java \cite{sedgewick2022introduction}, Python \cite{downey2015think}, and Pharo \cite{black2018pharo}. For each textbook, we built a \ac{RAG} system and measured its adherence to the textbook using established metrics \cite{DBLP:conf/emnlp/MinKLLYKIZH23,augenstein2023factuality,augenstein2024factuality,chang2024survey,spataru2024know} on a dataset of the 90 most popular questions about the textbooks' content, sourced from Stack Overflow. However, our findings indicate that all tested RAG-enhanced LLMs exhibit medium-to-low adherence to the reference textbook content (see Section \ref{sec:rq1}), despite being prompted to answer by paraphrasing the textbook.

We then ask how source faithfulness can be improved without sacrificing explanation quality.
To motivate our approach, we draw on Achinstein's theory of explanation \cite{achinstein1985nature}, which characterizes explaining as an illocutionary form of question answering: an explainer responds not only to the explicit query, but also to the implicit follow-up questions that a reasonable explainee would need resolved for understanding \cite{amsdottorato10943}.

This observation has a direct implication for retrieval-augmented systems. In practice, \acp{LLM} often behave \emph{as if} they were performing illocutionary macro-planning: they add background, definitions, and intermediate steps that help make an answer intelligible. We use \emph{macro-planning} here as a descriptive metaphor for this global content-selection behavior, not as a claim that transformers contain a discrete planning module. However, standard \ac{RAG} pipelines typically retrieve evidence only for the main question. When the model fills the remaining explanatory gaps, it must rely on prior knowledge that may be correct but is not necessarily aligned with (or traceable to) the reference textbook, thereby reducing source faithfulness.
From an XAI perspective, this reframes faithfulness as a property of the \emph{entire} explanatory macro-plan, not only of the final answer span.

To align this illocutionary macro-plan with the evidence source, we introduce \emph{chain-of-illocution prompting} (\texttt{CoI}): the system constructs a bank of meaningful, implicit explanatory questions from the textbook, selects a small set for each user query, uses them to drive retrieval, and then produces an explanation whose additional context is supported by the retrieved passages.
This yields our second and third research questions: \textbf{does illocutionary macro-planning improve source adherence in \ac{RAG}?} (\textbf{RQ2}) and \textbf{does chain-of-illocution compromise user satisfaction?} (\textbf{RQ3}).

To answer \textbf{RQ2}, we repeated the \textbf{RQ1} experiments with \texttt{CoI} added to the same RAG pipelines (cf. Section \ref{sec:rq2}). The results show that \texttt{CoI} significantly boosts source adherence for most evaluated RAG-enhanced LLMs, pushing these systems to favor textbook content over their internal knowledge, although the gains are smaller and not statistically significant for GPT-4o and Mixtral on some measures.
To answer \textbf{RQ3}, we also conducted a user study involving 220 recruited participants via Prolific (165 retained after filtering; cf. Section \ref{sec:rq3}). Participants were presented with Stack Overflow-approved correct explanations and then asked to assess their satisfaction with the AI-generated responses. Notably, the user study found no statistically significant decrease in satisfaction for \texttt{RAG+CoI} compared to \texttt{RAG}.

\paragraph{Contributions.}
We position source-faithful \ac{RAG} explanations as an XAI problem: explanations should be not only helpful but also \emph{traceable} to an explicit evidence source (here, a textbook).
Our contributions are:
\begin{itemize}
    \item \textbf{Theory:} We adapt Achinstein's illocutionary theory of explanation to retrieval-augmented settings, and propose \emph{illocutionary macro-planning} as a descriptive design principle for source-faithful explanations.
    \item \textbf{Method:} We introduce \emph{chain-of-illocution prompting} (\texttt{RAG+CoI}), a retrieval-driven prompting strategy that targets traceability by grounding implicit explanatory questions in the evidence source.
    \item \textbf{Evidence:} We benchmark six \acp{LLM} on 90 real-world Stack Overflow questions spanning three programming textbooks, quantify source faithfulness using FActScore and related metrics, and complement this with a user study with 150+ participants.
    \item \textbf{Artifacts:} We release the tools and data used in our experiments to support replication and follow-up work: \href{https://github.com/Francesco-Sovrano/chain-of-illocution}{https://github.com/Francesco-Sovrano/chain-of-illocution}
\end{itemize}

\section{Related Work \& Background} \label{sec:related_work} \label{sec:background}

\paragraph{RAG and Misinformation.}
LLMs can spread misinformation \cite{yang2024crag,vu2023freshllms,sun2024head}. In some cases, GenAI outputs are only 15\% accurate for rapidly evolving topics \cite{vu2023freshllms}. To address this, \ac{RAG} systems combine a retrieval mechanism that fetches relevant data from a source text (or database) with a \ac{GenAI} model that, guided by a task-specific prompt, uses this information to produce more accurate, contextually relevant responses. 
However, RAG systems still struggle with factuality and adherence to source texts \cite{yang2024crag,DBLP:conf/emnlp/MinKLLYKIZH23}. \citet{yang2024crag} show that while advanced LLMs achieve only 34\% factuality on the CRAG benchmark, simple RAG integration improves it to just 44\%. To standardize evaluation, \citet{DBLP:conf/emnlp/MinKLLYKIZH23} introduced FActScore (or \textit{source adherence precision}), which quantifies the percentage of atomic facts (subject-predicate-object triplets or clauses) in generated text supported by reliable sources, revealing that commercially available LLMs exhibit poor source adherence, with FActScores typically well below 70\%.
Our work builds on this evaluation perspective, but adapts it to textbook-grounded explanatory answers using clause-level semantic matching.

\paragraph{Micro- vs Macro-Planning.} 
The traditional distinction between micro- and macro-planning in natural language generation \cite{dale2000handbook} is key to understanding both misinformation challenges and GenAI outputs. Originating in psycholinguistics, \textit{macro-planning} sets the overall content and structure of a text, while \textit{micro-planning} handles detailed sentence organization \cite{levelt1999producing}. We use this distinction as an analytic vocabulary for the kinds of content decisions an \ac{LLM} makes while explaining, not as a claim that today's transformers contain an explicit macro-planning component. In GenAI, several studies \cite{puduppully2021data,kondadadi2013statistical,dale2000handbook} have explored neural models that separate macro-planning from surface realization. Most closely related, \citet{sovrano2024improve} extracted topic-related questions from textbook paragraphs to improve the explanatory power of an intelligent textbook. Our work builds on that idea, but moves it into a retrieval-time prompting pipeline for \ac{RAG}, evaluates it across multiple \ac{LLM} families, and focuses explicitly on source adherence rather than only explanatory richness.

\paragraph{Theories of explanation and traceability.}
Classic work in the philosophy and psychology of explanation treats explaining as a structured, audience-directed act rather than a mere statement of facts \cite{achinstein1985nature,holland1986induction,sellars1963philosophy,mayes2005theories}.
In particular, illocutionary accounts view explanations as answering an explicit question \emph{and} a set of implicit, context-building questions that make the answer intelligible \cite{sovrano2023perlocution}.
In \ac{RAG}, this perspective suggests an XAI design goal: the explanation's macro-structure should expose (and ground) these implicit questions so that each step remains traceable to retrieved evidence (source faithfulness), rather than being filled in from the model's prior knowledge.

\paragraph{Prompt Engineering Techniques.}
\citet{yang2024crag} highlight how prompt engineering enhances \acp{LLM} performance by using task-specific instructions without altering core parameters. For instance, chain-of-thought prompting \cite{vu2023freshllms} shows that adding a phrase like \quotes{Let's reason step by step} can elicit more reasoning in large language models. In \ac{RAG}, prompt techniques such as multi-query rewriting improve retrieval relevance and accuracy \cite{kostric2024surprisingly}, while research \cite{yu2020few,elgohary2019can} demonstrates that rewriting questions to be self-contained aids retrieval. Closest on the systems side are decomposition-based or interleaved retrieval approaches such as DSP/DSPy \cite{khattab2022demonstrate,khattab2024dspy}, Self-Ask \cite{press2023selfask}, and IRCoT \cite{trivedi2023ircot}, all of which break complex questions into intermediate steps or sub-questions. Our work is complementary but different in target: CoI does not primarily decompose a problem to improve answer accuracy on multi-hop QA, but to retrieve evidence for the \emph{implicit explanatory questions} that make an answer understandable and source-traceable.

Building on these insights, our work introduces chain-of-illocution prompting as a query-expansion strategy inspired by \citet{sovrano2024improve}. Rather than positing a literal planning module inside the model, CoI constrains the content-selection behavior that would otherwise be left implicit in generation by retrieving support for a small evidence-backed explanatory scaffold.

Prompt engineering can also be used not only to improve retrieval, but to increase the reliability of generation itself. Chain-of-Verification (CoVe) \cite{dhuliawala2024chain} is a prompt-based reliability technique that operates \emph{after} drafting: the model produces an initial answer, then generates verification questions, answers them, and revises the draft to address unsupported or incorrect claims. While CoVe strengthens responses through post-hoc self-critique, our chain-of-illocution (CoI) targets reliability \emph{upstream} by shaping the explanation \emph{before} generation. Consider the question \quotes{What is dependency injection?}. A CoVe-style workflow first drafts an answer, then checks claims such as whether dependencies are supplied from the outside and why this improves modularity, before revising the draft. CoI instead starts by retrieving evidence for implicit questions such as \quotes{What is a dependency?}, \quotes{What is the Dependency Inversion Principle?}, and \quotes{Why inject a dependency instead of instantiating it directly?}, and only then generates the explanation. Thus, CoI primarily improves the \emph{structure} and \emph{evidence-grounding} of explanations, whereas CoVe improves the \emph{correctness} of already-generated content. The two approaches are complementary: CoI can retrieve a faithful explanatory scaffold, and CoVe can subsequently verify and refine any residual unsupported claims.

This paper's evaluation perspective is also related to source-attribution work. ALCE \cite{gao2023alce} benchmarks end-to-end answer generation with citations, and the AIS framework \cite{rashkin2023ais} evaluates whether generated content is attributable to identified sources. Our notion of source adherence is aligned with this line of work, but tailored to single-source explanatory answers in \ac{RAG}, where the key question is not only whether a citation can be attached, but whether the explanatory content itself stays close to the authoritative textbook.

\section{Experiment Data: Evidence Sources (Textbooks) and Questions} \label{sec:book_n_questions}

\paragraph{Textbooks.} For reproducibility, we selected open-access textbooks. We chose two on mainstream languages (Java \cite{sedgewick2022introduction} and Python \cite{downey2015think}) and one on a less common programming language (Pharo \cite{black2018pharo}). When multiple options were available, we chose the most comprehensive textbook to ensure broader content coverage. Table \ref{tab:textbook_stats} summarizes each textbook's statistics.

\begin{table}[t]
    \centering
        \begin{tabular}{lrrrcl} 
            \hline
            \textbf{Tag} & \textbf{Textbook} & \textbf{\#Pages} & \textbf{\#Words} & \textbf{Year} & \textbf{Copyright} \\
            \hline
            java & \cite{sedgewick2022introduction} & 773 & 350'935 & 2022 & CC BY-NC-SA 4.0 \\
            python & \cite{downey2015think} & 244 & 68'202 & 2015 & CC BY-NC 3.0 \\
            pharo & \cite{black2018pharo} & 376 & 90'572 & 2018 & CC BY-SA 3.0 \\
            \hline
        \end{tabular}
    \caption{Statistics of Selected Textbooks.} \label{tab:textbook_stats}
\end{table}

\paragraph{Questions.} Our question selection criteria aimed to evaluate RAG's textbook adherence using common, everyday challenges that students face, not artificially hard questions for RAG systems. Hence, we focused on popular, widely viewed questions to ensure practical relevance. Given our focus on programming education, we could rely on Stack Overflow, a Q\&A platform that provides statistics on question popularity, to identify the most popular questions related to the textbooks' content. For each textbook, we identified a corresponding Stack Overflow tag (\texttt{java}, \texttt{python}, \texttt{pharo}) and used a SQL query on the Stack Exchange Data Explorer to extract up to the 200 most viewed questions per tag. We filtered out questions without an approved answer (needed to assess AI-generated explanations) and manually removed those outside the book's scope (e.g., installation issues, non-standard libraries). This resulted in excluding 8 questions for \texttt{java}, 9 for \texttt{python}, and 3 for \texttt{pharo}. We then selected the top 30 questions per tag, totaling 90 questions.

An \textit{a priori} power analysis confirmed that this dataset exceeds the 74-question minimum needed for the paired RQ2 comparisons (assuming a small-to-moderate effect size of Cohen's $d_z = 0.3$, $1-\beta \geq 0.8$, and $\alpha \leq 0.05$)~\cite{lakens2022sample}. Because \textbf{RQ2} compares \texttt{RAG+CoI} and \texttt{RAG} on the same set of questions, the relevant inferential setting is within-question paired testing rather than independent-samples testing. Accordingly, the dataset is sufficiently large to provide adequate power for the paired analyses reported in Section~\ref{sec:rq2}.
%

\begin{figure}
    \centering
    \includegraphics[width=\linewidth]{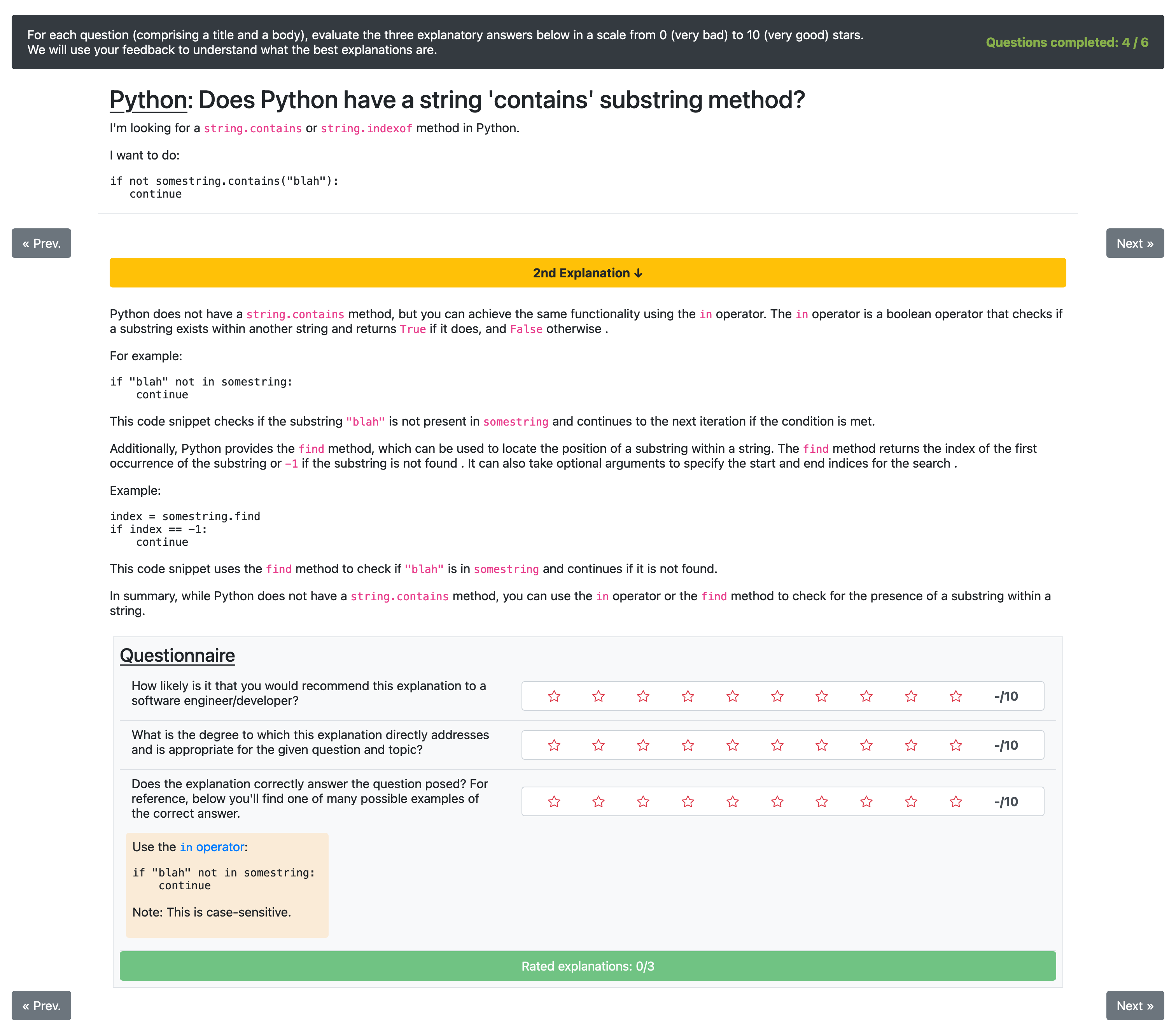}
    \caption{Study interface with an explanation generated via chain-of-illocution.}
    \label{fig:study_interface}
\end{figure}

The complete list of questions used in this study is available in the online replication package. Table \ref{table:questions_stats} also presents some statistics for these questions. 
An example of a complete question, including both the title and body, is shown in Figure \ref{fig:study_interface}. 
Examples of questions are: \quotes{What is dependency injection?}; \quotes{How do I convert a String to an int in Java?}.

\begin{table}
    \centering
    {%
        \begin{tabular}{lr|r|r} 
        \hline
        \textbf{Tag}    & \multicolumn{1}{l|}{\begin{tabular}[c]{@{}r@{}}\textbf{Question Size}\\\textbf{Avg.}\end{tabular}} & \multicolumn{1}{l|}{\begin{tabular}[c]{@{}r@{}}\textbf{Body Size}\\\textbf{Avg.}\end{tabular}} & \multicolumn{1}{l}{\begin{tabular}[c]{@{}r@{}}\textbf{Approved}\\\textbf{Answer Size}\\\textbf{Avg.}\end{tabular}}  \\ 
        \hline
        java            & 8.67                                                                                                             & 53.3                                                                                                         & 287.47                                                                                                                               \\
        python          & 9.37                                                                                                             & 51.83                                                                                                        & 265.37                                                                                                                               \\
        pharo           & 9.57                                                                                                             & 95.1                                                                                                         & 181.57                                                                                                                               \\
        \hline
        \end{tabular}
    }
    \caption{Statistics of the selected Stack Overflow questions. Lengths are approximate whitespace-token counts computed on the raw exported Stack Overflow fields.}
    \label{table:questions_stats}
\end{table}

\section{RQ1: Source Faithfulness Benchmark (Textbook Adherence)} \label{sec:rq1}

\textbf{RQ1:} \textit{To what extent do RAG systems adhere to their reference textbook?}

\paragraph{Methodology.} To answer \textbf{RQ1} and evaluate the source adherence of widely used RAG-enhanced \acp{LLM}, we implemented a standard \ac{RAG} pipeline for each textbook in Table \ref{tab:textbook_stats}, varying the LLM. For each textbook, we posed to its RAG system only the questions corresponding to that textbook (e.g., only the questions tagged as \texttt{java} for the Java textbook) and examined how the FActScore \cite{DBLP:conf/emnlp/MinKLLYKIZH23}, along with other secondary metrics detailed below, varied for each LLM. Notably, FActScore and related factual-precision metrics are widely used to measure source adherence precision \cite{augenstein2023factuality,augenstein2024factuality,dhuliawala2023chain,chang2024survey,spataru2024know}. In our study, we use an adapted FActScore-style adherence metric tailored to textbook-grounded explanations.

As a reference baseline, we also compared each RAG pipeline with its corresponding bare LLM operating without retrieval augmentation and relying solely on direct questioning. Finally, despite LLMs' non-determinism \cite{DBLP:journals/corr/abs-2307-09009}, we reduced output variability by using a highly constrained decoding configuration in the OpenAI calls (temperature=0.5, top-p=0)\footnote{Because the precise runtime semantics of top-p=0 are API-dependent, we report the exact parameterization rather than characterizing it as strictly greedy decoding.}. Lower temperatures (0--0.5) reduce randomness, while lower top-p values restrict diversity.

The \texttt{GenAI} explanations were generated using a prompt analogous to that used for \texttt{RAG}. In the released code, this prompt is passed as a user message rather than as a chat ``system'' message. The question-answering template is:
\begin{quote}\itshape
Provide a detailed, concise, pertinent, and coherent explanatory answer to the question below. Provide examples if needed.\\
\\
Question:\\
\#\{topic\}\\
\{body\}
\end{quote}
Notably, this prompt does not reference any retrieved chunks or textbooks, as \texttt{GenAI} does not employ an information retrieval system.

\paragraph{RAG Implementation.} We implemented a standard dense-retrieval \ac{RAG} pipeline (similar to OpenAI's published guidelines) for our information retrieval system. This approach encodes queries and document chunks into high-dimensional vectors using OpenAI's \texttt{text-embedding-3-large} model, and identifies relevant chunks via cosine similarity. For the generative component, we evaluated six widely used LLMs (see Table \ref{table:model_specifications} for details), chosen from three widely used model families: GPT (GPT-3.5-turbo and GPT-4o), Llama 3 (8B and 70B models), and Mistral (7B and 8×7B models). We used the equivalent of OpenAI's Assistants RAG system for textbook chunking, with 150-token chunks (minimum 100) and a 75-token overlap. The maximum number \( k \) of retrieved chunks was set to 10, as this setting allowed for query expansion, given the context window size limitations of the \acp{LLM}, with little observed performance difference from other settings. 

Finally, the prompt used by RAG to integrate the retrieved text chunks was designed to encourage the LLM to produce detailed, concise, pertinent, and coherent explanatory answers. Emphasizing conciseness was essential because \citet{kabir2024stack} noted that verbosity is common in ChatGPT outputs. Moreover, the prompt requires each RAG statement to reference source textbook page(s) and provide examples when needed, as examples are critical for high-quality programming explanations \cite{nasehi2012makes}. The exact prompt released in the replication package for the question-answering setting is:
\begin{quote}\itshape
    Sift through the text chunks provided (extracted from the textbook ``\{textbook\}'') and combine the most relevant ones into a detailed, concise, pertinent, and coherent explanatory answer to the question below. Every statement must contain a reference to the source textbook page(s). Provide examples if needed.\\
\\
    Question:\\
    \#\{topic\}\\
    \{body\}\\
\\
    Text chunks:\\
    \{contents\}
\end{quote}


\begin{table}[t]
    \centering
    {%
        \begin{tabular}{lcrcr} 
        \hline
        \textbf{Model} & \begin{tabular}[c]{@{}c@{}}\textbf{Know. }\\\textbf{Cut-off}\end{tabular} & \multicolumn{1}{c}{\textbf{Size}} & \begin{tabular}[c]{@{}c@{}}\textbf{Open}\\\textbf{Source}\end{tabular} & \multicolumn{1}{c}{\begin{tabular}[c]{@{}c@{}}\textbf{Max. }\\\textbf{Tokens}\end{tabular}}  \\ 
        \hline
        gpt-3.5-turbo  & 2021-09                                                                   & -                                 & No                                                                     & 16k                                                                                          \\
        gpt-4o         & 2023-10                                                                   & -                                 & No                                                                     & 128k                                                                                         \\
        \hline
        Llama 3:8b-instruct    & 2023-12                                                                   & 8B                               & Yes                                                                    & 8k                                                                                           \\
        Llama 3:70b-instruct    & 2023-12                                                                   & 70B                               & Yes                                                                    & 8k                                                                                           \\
        \hline
        mistral:7b-instruct        & 2021-08                                                                   & 7B                                & Yes                                                                    & 32k                                                                                          \\
        mixtral:8x7b-instruct   & 2023-12                                                                   & 45B                               & Yes                                                                    & 32k                                                                                          \\
        \hline
        \end{tabular}
    }
    \caption{Statistics of selected \acp{LLM}.}
    \label{table:model_specifications}
\end{table}

\paragraph{Metrics Implementation.}
We evaluate source adherence using an adapted FActScore-style metric inspired by \citet{DBLP:conf/emnlp/MinKLLYKIZH23}, but operationalized for textbook-grounded explanatory answers.
First, we extract grammatical clauses (i.e., subject, predicate, and object) from each explanation. Then, using the \texttt{text-embedding-3-large} model, these clauses are embedded and matched to similar clauses from the textbook. Finally, \textit{source adherence} is determined by the ratio of AI-generated clauses with a similarity score above \( t = 0.7 \) (i.e., an AI-generated clause is considered equivalent to a textbook clause only if its similarity is at least 70\%), yielding a value in $[0,1]$. 

To summarize, our implementation of FActScore uses embedding-based clause matching and thresholding, rather than the original evaluation protocol in \cite{DBLP:conf/emnlp/MinKLLYKIZH23}.
For brevity, we report FActScores only at \( t = 0.7 \), though we examined other values. For \( t \ll \frac{2}{3} \), any AI-generated clause mentioning at least one subject, predicate, or object from the textbook is deemed adherent, resulting in an unacceptable FActScore of 100\% for details outside the textbook scope. Conversely, \( t \geq 0.9 \) results in numerous false negatives and FActScores near 0\% since not only must the subject, predicate, and object match, but they must also use identical wording. For \( \frac{2}{3} < t < 0.9 \), a clause is considered adherent if its subject, predicate, and object (each weighted $\frac{1}{3}$) match the textbook version, even when paraphrased. Empirical observations also indicate that thresholds between 0.6 and 0.8 best balance false positives and negatives.

As an additional metric, we also compute the \textit{mean semantic similarity} between AI-generated and textbook clauses (equivalent to a FActScore with \( t = 0 \)) where a value of 1 indicates nearly identical clauses. This metric complements the FActScore by focusing on semantic similarity, while the FActScore measures source adherence. Moreover, since FActScore assesses adherence precision (a ratio) and not the absolute number of adherent clauses \cite{DBLP:conf/emnlp/MinKLLYKIZH23}, thus we also report the \textit{number of adherent clauses} for a more comprehensive evaluation.

\paragraph{Results.} Non-RAG GenAI models have a median textbook adherence precision of 0\%, even though they answer questions correctly. In contrast, RAG models show low median source adherence: GPT-4o (22.2\%), Llama 3-70B (33.3\%), Llama 3-8B (33.3\%), Mistral (23.6\%), Mixtral (29.3\%), and GPT-3.5-turbo (40\%). However, their mean semantic similarity with textbook content stays above 60\% (up to 67.8\% for GPT-3.5-turbo), implying that while the generated explanations have high cosine similarity (which does not necessarily mean \emph{similar meaning}), they often do not resemble the source texts (low source adherence). Finally, most RAG systems generate explanations with a median of 2 to 3 adherent clauses. The explanations typically range from 212 to 295 words, likely due to the prompt requiring conciseness. 

For more details, see Figure \ref{fig:factuality_study_results} in Section \ref{sec:rq2}, summarizing findings for both \textbf{RQ1} and \textbf{RQ2}.

\paragraph{Discussion.} Although the prompt explicitly told the RAG systems to reference the textbooks and combine the relevant chunks provided, we observed that the models often relied on prior knowledge, introducing information not present in the context window. This tendency is especially noticeable when explanatory examples are provided as part of the explanation by the LLM, though it is not limited to those cases. For instance, in response to \quotes{In Pharo/Smalltalk: How to read a file with a specific encoding?}, the GPT-4o-based RAG recommends using a \texttt{FileStream} with a text converter, making the example of the \texttt{UTF8TextConverter} for UTF-8. However, the Pharo book \cite{black2018pharo} never mentions \texttt{UTF8TextConverter}.

\begin{finding}{Summary of RQ1}
    All tested RAG-enhanced \acp{LLM} have median \textit{source adherence} below 50\%.
    Adherence is often lower for larger models, though the trend is not monotonic across model families.
\end{finding}

\section{RQ2: Illocutionary Macro-Planning for Source-Faithful Explanations} \label{sec:rq2}

\begin{figure*}
    \centering
    \includegraphics[width=\linewidth]{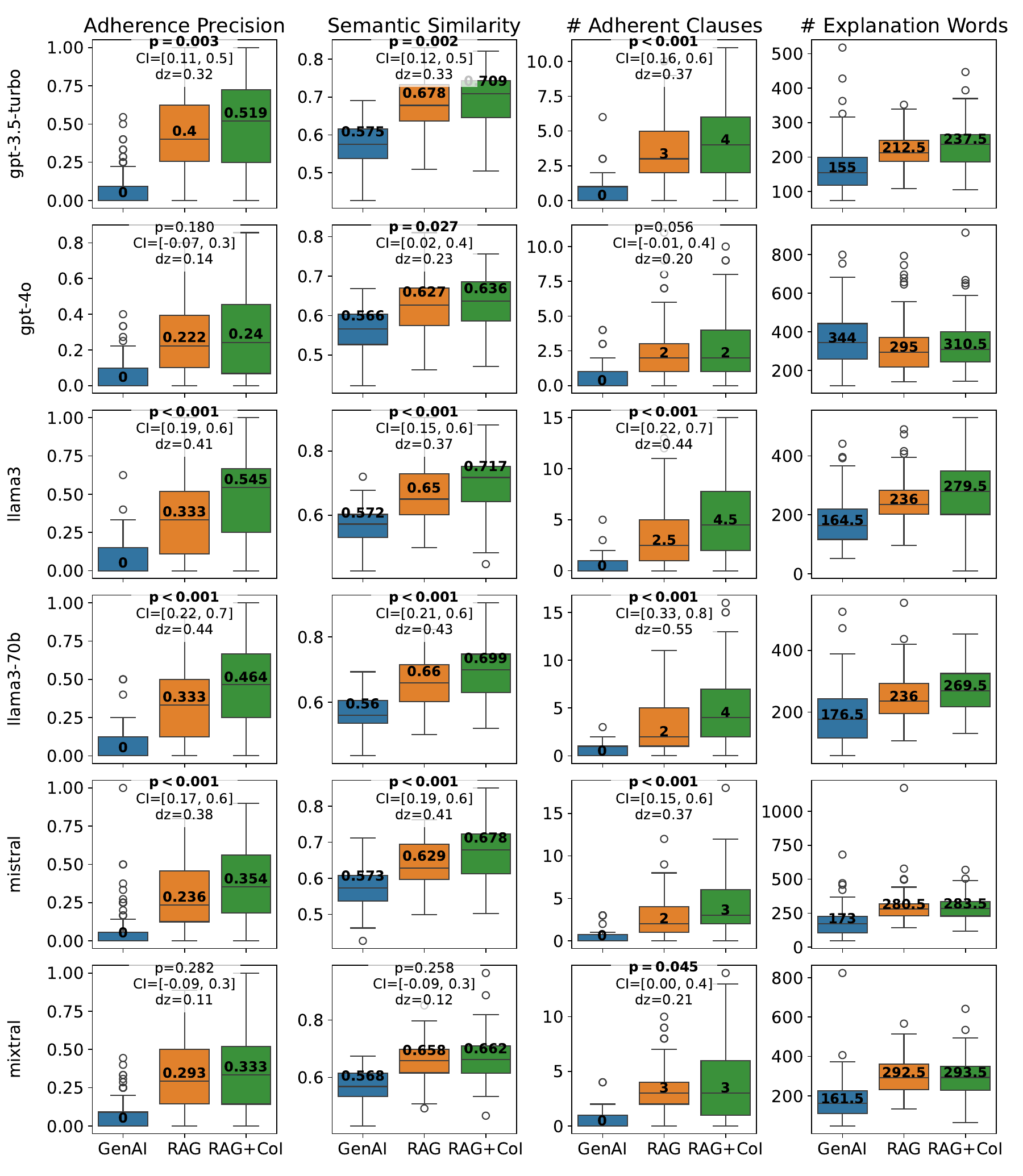}
    \caption{Source faithfulness across six \acp{LLM}, with one-sided p-values (significant in bold), Cohen's dz effect sizes, and 95\% confidence intervals (CI).}
    \label{fig:factuality_study_results}
\end{figure*}

\textbf{RQ2:} \textit{How can source faithfulness (textbook adherence) in RAG be improved without training?}

\begin{principle}{Principle 1 (Illocutionary grounding)}
For an explanation to be \emph{source-faithful} and thus scrutable, the system should ground not only the final answer but also the implicit context-building questions that constitute the explanation's illocutionary macro-plan.
Operationally, this means \textit{(i)} eliciting a small set of implicit guiding questions, \textit{(ii)} retrieving evidence for each, and \textit{(iii)} generating the explanation using only what is supported by the retrieved evidence so that claims remain traceable to the source.
\end{principle}

\paragraph{Methodology.} Based on the \textbf{RQ1} results, we see that \acp{LLM} often add background information that helps an explanation read well but is not always supported by the retrieved textbook passages. An explanation is not merely an answer to the main question; it must also address implicit questions to help non-experts grasp the surrounding context \cite{achinstein1985nature}. This act of answering implicit questions is known as \textit{illocution} \cite{sovrano2023objective,amsdottorato10943}. Our goal is therefore to retrieve evidence not only for the main question, but also for a compact evidence-backed explanatory scaffold of relevant implicit questions.

\begin{figure*}
    \centering
    \includegraphics[width=.85\linewidth]{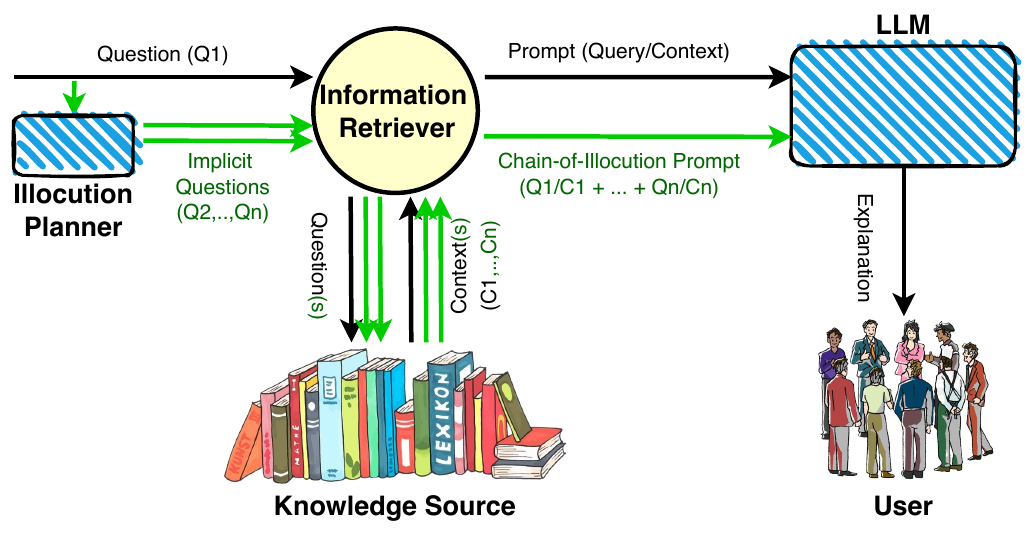}
    \caption{Chain-of-illocution pipeline. Black arrows denote the standard RAG flow from the user question to retrieval, prompting, and final explanation. Green arrows denote the CoI additions: an illocution planner that proposes implicit explanatory questions, per-question retrieval from the knowledge source, and the concatenation of question--context pairs $(Q_1,C_1), \ldots, (Q_n,C_n)$ into the final prompt. The multiple green arrows between the retriever and the knowledge source indicate that evidence is fetched separately for multiple implicit questions, not only for the user's original question.}
    \label{fig:chain_of_illocution_diagram}
\end{figure*}

This can be achieved via prompt engineering, using a query expansion approach similar to \citet{kostric2024surprisingly} and the macro-plan for explanatory illocution of \citet{sovrano2024improve}. We call this the \textit{chain-of-illocution} strategy. Figure \ref{fig:chain_of_illocution_diagram} previews the pipeline. To test it, we repeated the \textbf{RQ1} experiment using the same answer-generation prompt, with and without the chain-of-illocution query expansion (\texttt{RAG+CoI}). We then conducted an ablation study comparing the adherence precision of \texttt{RAG+CoI} to the baseline (\texttt{RAG}). Following a Shapiro-Wilk normality test, non-parametric Wilcoxon signed-rank tests were used for FActScore and number of adherent clauses (since they were not normally distributed), while paired t-tests were applied to mean semantic similarity. Unless otherwise stated, the reported p-values in this section are one-sided and aligned with the directional hypothesis that \texttt{RAG+CoI} improves on \texttt{RAG}.

\paragraph{Implementation.} To implement chain-of-illocution, we first construct an offline bank of implicit questions that differ from the main question and provide additional explanatory context. Instead of merely rewriting the original question with an LLM at inference time (see Section \ref{sec:background}), we extract these questions directly from the textbook content and later retrieve a small subset for each Stack Overflow question. Following the procedure in \cite{sovrano2024improve} with the Mistral model, we extract questions for each paragraph.

We initially tried a template-based method using fixed archetypal questions (e.g., \quotes{Why is \{X\}?}, \quotes{How is \{X\}?}), but it produced too many ungrammatical or irrelevant questions because the textbooks often do not provide the local context needed to instantiate those templates cleanly. In contrast, our dynamic, content-driven approach generates contextually relevant and adaptable questions. Using the LLM-based technique, we extracted 40'737 implicit questions for the \texttt{java} textbook, 9'879 for \texttt{python}, and 13'362 for \texttt{pharo}.

Specifically, we employed the Mistral model (see Table \ref{table:model_specifications}) with the following prompt (shown here in its English-language instantiation):
\begin{quote}\itshape
Analyse the English paragraph below to generate a comprehensive list of Q\&As in English, capturing: what, who, why, how, how much, where, when, who by, which, whose. Answers must succinctly reflect the paragraph's content without repeating the question's wording. Q\&As must use precise and direct language, avoiding vague terms and generalizations, clearly specifying the context and subjects involved without assuming prior knowledge.\\
\\
Example Paragraph: Alice, an experienced hiker, explores the Rocky Mountains despite rain. She packs her gear early in the morning.\\
\\
Expected Output:\\
- Who is Alice? An experienced hiker.\\
- What did Alice do? Explored the Rocky Mountains.\\
- Despite what did Alice decide to explore the Rocky Mountains? Rain.\\
- What did she pack? Gear.\\
- When did she pack? Early in the morning.\\
\\
Paragraph for Analysis:\\
\{sentence\}
\end{quote}

The CoI pipeline for each Stack Overflow question is:
\begin{enumerate}
    \item retrieve a candidate pool of \(M=25\) semantically related implicit questions from the textbook-derived question bank using \texttt{text-embedding-3-large} and cosine similarity;
    \item augment that pool with template-generated \quotes{What is \{X\}?} questions, where \(X\) is a subject or object label from the primary question;
    \item retrieve up to \(k=10\) textbook chunks for each candidate question and discard candidate questions for which no supporting chunk is found;
    \item if the same chunk is retrieved for multiple candidate questions, assign it only to the question with the highest similarity score;
    \item rank the remaining candidate questions by the similarity of their best supporting chunk and keep the top \(m=5\) implicit questions as the final explanatory scaffold; and
    \item concatenate the original question with the selected question--context pairs and ask the \ac{LLM} to generate the final explanation.
\end{enumerate}

We use \(M=25\) as an over-generation pool and keep \(m=5\) final implicit questions. This 5:1 ratio gives the planner enough redundancy to filter out unsupported or duplicate candidates while keeping the final expanded prompt within the context-window limits of the evaluated models (see Table \ref{table:model_specifications}). Among the template-generated questions, only the \quotes{what} questions consistently produced meaningful additions.

For example, the final five implicit questions retrieved for \quotes{What is dependency injection?} are:
\begin{inparaenum}[\itshape i\upshape)]
    \item \quotes{How should the high-level class interact with the low-level class as per the Dependency Inversion Principle?}
    \item \quotes{How can you apply the dependency inversion principle and open/closed principle?}
    \item \quotes{What is the effect of applying the dependency inversion principle?}
    \item \quotes{What is dependency in the context of classes?}
    \item \quotes{What is the Dependency Inversion Principle?}
\end{inparaenum}

The selected implicit questions, along with the primary one, are then sent to the information retriever to fetch up to $k$ chunks per question for the final prompt. The retrieved questions and chunks are concatenated with the original query and context as shown in Figure \ref{fig:chain_of_illocution_diagram}.

\paragraph{Results.} Figure \ref{fig:factuality_study_results} shows statistically significant one-sided improvements ($p \!<\! 0.05$; 95\% confidence intervals included) for \texttt{RAG+CoI} over \texttt{RAG} (and consequently also \texttt{GenAI}) across most tested models and metrics, with weaker or non-significant effects for GPT‑4o and Mixtral. The figure reports the raw p‑values; we then applied the Benjamini-Hochberg procedure \cite{benjamini1995controlling} to control the false discovery rate at 5\%, and every comparison with a raw $p \!\leq\! 0.027$ (i.e., all but one of those originally significant) remained significant, underscoring the robustness of these findings. The sole exception was Mixtral's $p\!=\!0.045$.
Figure \ref{fig:factuality_study_results} also shows that the \texttt{RAG+CoI} explanations include, on average, 1--2 more textbook-adherent clauses than those from \texttt{RAG}, despite similar explanation lengths. These gains in adherent clauses are supported by small-to-medium Cohen's dz effect sizes across models ($dz \!=\! 0.367$ for Mistral up to $dz \!=\! 0.550$ for Llama 3-70B) indicating consistent, statistically reliable increases in factuality. \texttt{RAG+CoI} also achieves higher mean semantic similarity across all models, though the Mixtral gain is very small and non-significant; improvements range from +0.6\% (Mixtral; $p \!=\! 0.282$, non-significant) to +10\% (Llama 3-8B; $dz \!=\! 0.367$; $p \!<\! 0.001$), with most models exhibiting moderate ($dz \!\approx\! 0.4$, except for GPT-4o) effect sizes that reflect small-to-medium gains. 

Adherence precision follows a similar pattern, improving by +8.1\% (GPT-4o; $p \!=\! 0.180$, non-significant) to +63.7\% (Llama 3-8B; $dz\!=\!0.436$; $p \!<\! 0.001$), again demonstrating small-to-medium effects that underline the robustness of the precision gains. Overall, \texttt{RAG+CoI} explanations are only 5.1\% more semantically similar but 34.08\% more adherent, on average, compared to \texttt{RAG}. Among the models, Llama 3-8B achieves the highest adherence precision (0.545; $dz \!=\! 0.408$; $p \!<\! 0.001$), followed by GPT-3.5-turbo (0.519; $dz \!=\! 0.317$; $p \!=\! 0.003$) and Llama 3-70B (0.464; $dz \!=\! 0.436$; $p \!<\! 0.001$), while Mistral (0.354; $dz \!=\! 0.382$; $p \!<\! 0.001$) and GPT-4o (0.333; $p\!=\!0.180$, non-significant) remain below 0.4, underscoring the model-specific variability in how strongly \texttt{RAG+CoI} boosts factual alignment. 

Notably, given the \textit{a priori} power analysis presented in Section \ref{sec:book_n_questions}, the paired \textbf{RQ2} comparisons with observed \(dz \ge 0.3\) are expected to have adequate power (\(1-\beta \ge 0.8\)). This applies to the question-level \texttt{RAG+CoI} vs.\ \texttt{RAG} analyses, and should not be read as a blanket power claim for all other tests reported in the paper.

\paragraph{Discussion.} Larger models (GPT-4o, Mixtral, Llama 3-70B) tend to have lower adherence precision than smaller ones (GPT-3.5-turbo, Mistral, Llama 3-8B), though the trend is not monotonic across model families. However, while smaller models follow the prompt more strictly, they are more vulnerable to retrieval errors. For example, when asked about the difference between \texttt{self} and \texttt{yourself} in Smalltalk, the retrieval system missed the scarce reference to \texttt{yourself} in the Pharo textbook \cite{black2018pharo}, leading smaller models to wrongly claim that the operator does not exist. Larger models usually avoided this mistake by relying on prior knowledge.

These gains should nevertheless be interpreted cautiously. Absolute adherence remains moderate even after CoI: the best median adherence precision is still only around 0.55, while several models remain between roughly 0.33 and 0.46. Moreover, the improvements are weak or non-significant for GPT-4o ($p\!=\!0.180$) and for Mixtral on adherence precision / semantic similarity ($p\!=\!0.282$), indicating that CoI is not a universal fix and that its benefits depend on how strongly a model follows retrieved evidence instead of falling back on prior knowledge.
\begin{finding}{Summary of RQ2}
    Chain-of-illocution prompting improves source faithfulness (textbook adherence) precision by an average of 34.08\%, with a maximum median increase of 63.66\%.
\end{finding}

\section{RQ3: User Study} \label{sec:rq3}

\textbf{RQ3:} \textit{Does chain-of-illocution compromise user satisfaction?}
\paragraph{Methodology.} We conducted a user study to test whether chain-of-illocution compromises user satisfaction. We compared \texttt{RAG+CoI} with \texttt{RAG}, and treated perceived correctness and relevance as secondary outcomes alongside satisfaction. We expected either no significant difference or an improvement in these user-facing judgments.
We ran a mixed-design user study focusing on GPT-4o and GPT-3.5-turbo to control costs (about £110 per model-specific study; about £220 total). Model (GPT-4o vs. GPT-3.5-turbo) was treated as a between-subjects factor, while system condition (GenAI, RAG, and RAG+CoI) was treated as a within-subjects factor: within each model-specific study, participants compared explanations from all three systems for each question.
We selected these two GPT models because GPT-4o showed substantially lower adherence precision than GPT-3.5-turbo in \textbf{RQ1}, making it the least source-adherent GPT model we tested, whereas GPT-3.5-turbo was the most source-adherent GPT model. We recruited about 110 participants per model (220 total) via Prolific, selecting for English fluency, programming knowledge (e.g., Java, Python), software development experience, and an approval rate of 80\%+.

\paragraph{Study Design.} Participants compared explanations from three systems (\texttt{RAG}, \texttt{RAG+CoI}, and \texttt{GenAI}) for each question. They were also given the Stack Overflow-approved correct explanation as the ground truth. To simplify the process, source-reference markers were removed automatically before display, eliminating the need to provide the textbook. In the released interface code, this cleanup is implemented with regular expressions that remove bracketed citation spans and parenthesized source annotations. 

Participants evaluated each explanation on satisfaction (\quotes{How likely are you to recommend this explanation?}), relevance (\quotes{To what extent does this explanation directly address and appropriately cover the given question and topic?}), and perceived correctness (\quotes{Does the explanation correctly answer the question posed? For reference, below is one of many possible correct answers: \{accepted\_answer\}}).
All questions were rated on a Likert scale (0 = very low/bad to 10 = very high/well). Satisfaction was measured using the Net Promoter Score template \cite{fisher2019good,baquero2022net}, though we retained the original 0--10 scores for analysis. Relevance and correctness metrics were adapted from \cite{es2024ragas}. A Likert scale better reflects perceived correctness and uncertainty than a binary approach, and provides greater statistical power.

Each participant was paid 1£ for an experiment reported to take 10 minutes. Participants evaluated explanations for 6 out of the 90 Stack Overflow questions (18 explanation ratings in total, since each question was shown with three systems). The question pool was drawn from the three textbook-related tags (\texttt{java}, \texttt{python}, \texttt{pharo}), and the assignment was balanced so that the explanations for each question were eventually reviewed by at least 3 different participants.
%
After the study, we excluded participants who spent less than 6 minutes in total (i.e., less than about 1 minute per question, or about 20 seconds per explanation) or rated every explanation 10 stars. This filtering retained 90 for GPT-3.5-turbo and 75 for GPT-4o.

\begin{figure*}
    \centering
    \includegraphics[width=.85\linewidth]{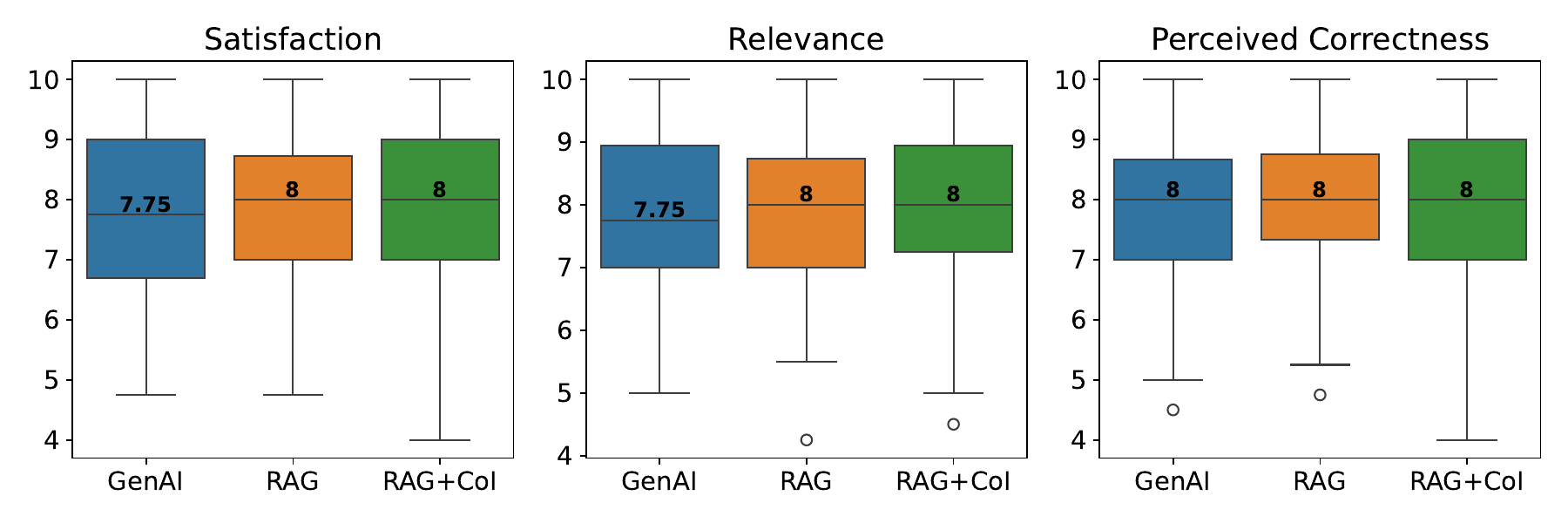}
    \caption{User study results for GPT-3.5-turbo.}
    \label{fig:user_study_results_gpt-3.5-turbo}
\end{figure*}

\begin{figure*}
    \centering
    \includegraphics[width=.85\linewidth]{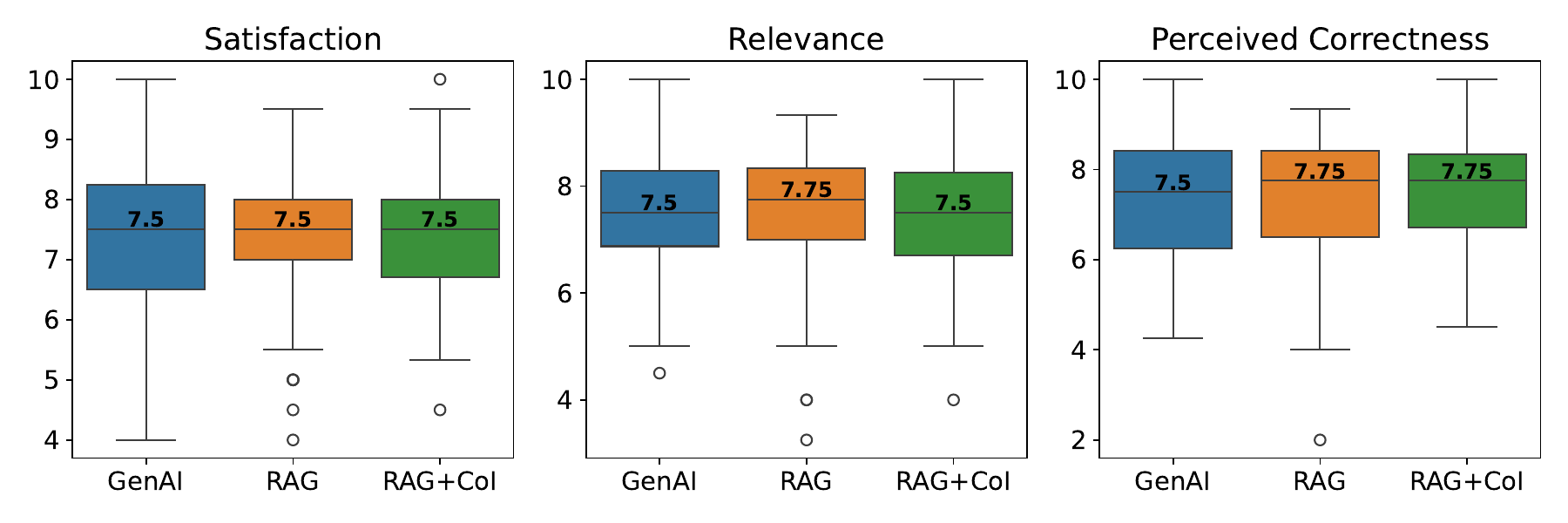}
    \caption{User study results for GPT-4o.}
    \label{fig:user_study_results_gpt-4o}
\end{figure*}




\paragraph{Results.} We averaged participant scores and examined their distributions (see Figures \ref{fig:user_study_results_gpt-3.5-turbo} and \ref{fig:user_study_results_gpt-4o}). One-sided Wilcoxon signed-rank tests did not reveal significant decreases in satisfaction, relevance, or perceived correctness for \texttt{RAG+CoI} relative to \texttt{RAG}, indicating that \textbf{chain-of-illocution prompting does not harm these metrics} in this study. 
We also performed one-sided Mann-Whitney U tests to assess whether the lower textbook adherence observed with the larger LLM (GPT-4o) negatively impacted the metrics under scrutiny. 

The tests revealed that GPT-3.5-turbo's greater adherence was associated (see discussion below) with significantly ($p<0.05$, one-sided) higher satisfaction ($U=3973.5$, $p=0.024$ for \texttt{RAG}; $U=4015.5$, $p=0.017$ for \texttt{RAG+CoI}), significantly improved perceived correctness under \texttt{RAG} ($U=4034.5$, $p=0.015$), and significantly greater relevance under \texttt{RAG+CoI} ($U=3947.0$, $p=0.03$). Differences in relevance for \texttt{RAG} ($p=0.11$) and correctness for \texttt{RAG+CoI} ($p=0.069$) were not statistically significant. 
Even when adjusting the p-values above via Benjamini-Hochberg correction \cite{benjamini1995controlling}, the results remain significant.

Finally, the Cohen's $d$ values for the Mann-Whitney U tests above are between 0.19 to 0.34, indicating small effect sizes by conventional benchmarks. This suggests that although the differences are consistent and statistically reliable, their magnitude remains modest.

\paragraph{Discussion.} Explanations generated with \texttt{RAG} and \texttt{RAG+CoI} on GPT-3.5-Turbo received the highest median ratings for satisfaction, relevance, and perceived correctness. Notably, neither system's scores are significantly higher (only slightly) than that of \texttt{GenAI} on either GPT-3.5-Turbo or GPT-4o. Moreover, \texttt{RAG} and \texttt{RAG+CoI} do not differ visibly from each other, which is fine.  
Indeed, although one might expect that greater (textbook) adherence would correspond to higher satisfaction, relevance, or perceived correctness, this is not the case, for two reasons.  

The first reason is that the questions we considered are, by design, among the most popular on the web (i.e., from Stack Overflow). A \texttt{GenAI} model like ChatGPT clearly had access to them and learned to answer them correctly, as shown by the results. Additionally, ChatGPT is also very fluent, so it makes sense that not only the perceived correctness scores are high, but also satisfaction and relevance. \texttt{RAG} is not a newly trained model, it is simply \texttt{GenAI} with access to external resources, which in our case are the textbooks. Hence, \texttt{RAG} inherits all the knowledge and fluency of \texttt{GenAI} by design. The same goes for \texttt{RAG+CoI}.

The second important reason is that satisfaction, relevance, and perceived correctness are not measures of adherence. Higher adherence does not imply higher satisfaction, relevance, or perceived correctness. We simply did not know whether it implied lower scores instead, something our user study shows is not the case.
Then why do the lower adherence scores of GPT-4o correlate with significantly lower satisfaction, relevance, and perceived correctness?

We believe this is because the baseline GPT-4o model has a less precise understanding of how to correctly answer the questions we considered, leading to slightly worse outputs overall. That, of course, can only be partly mitigated by \texttt{RAG} and \texttt{RAG+CoI}. Indeed, if we look at Figure \ref{fig:user_study_results_gpt-4o}, the improvement in median perceived correctness for both over \texttt{GenAI} is only 0.25 points. This is likely because if the underlying LLM \quotes{believes} a piece of context (even if retrieved from an authoritative source like a textbook) is not useful for the answer, it will not use it.  

We verified this through a qualitative error analysis of GPT-4o's outputs, which indicated that the larger model often struggles to incorporate missing context, something that appears to be the primary reason for the lower satisfaction and perceived correctness scores.

What really matters is that, overall, the chain-of-illocution strategy was not negatively received, indicating strong potential for real-world applications that require strict textbook adherence (which, of course, is not always a necessary feature in an educational context). 
Indeed, chain-of-illocution can significantly increase adherence without hindering user satisfaction, relevance, and perceived correctness. Conversely, when such adherence is not necessary, our results suggest that (at least for the topics examined) RAG is not needed as \texttt{GenAI} alone suffices. In other words, if an educator wants more control over the content these LLMs generate in an educational context, \texttt{RAG+CoI} is the strategy to use. 

\begin{finding}{Summary of RQ3}
    Chain-of-illocution prompting does not compromise user satisfaction, perceived correctness, and relevance. 
\end{finding}

\section{Threats to Validity} \label{sec:threats_to_validity}

\paragraph{Internal Validity.} The non-deterministic nature of LLMs may cause variability in responses, affecting our study's consistency \cite{DBLP:journals/corr/abs-2307-09009}. To address this, we averaged results from 90 questions per LLM and applied statistical analyses. Similarly, to counter variability in human evaluations, we recruited over 100 participants per LLM and averaged their ratings.

\paragraph{External Validity.} Our findings, based on specific LLMs, programming languages, and textbooks, might not generalize to other settings. To address this, we employed a diverse range of LLMs (both open- and closed-source) and over 90 questions across three topics. To enhance representativeness, we also chose widely-used textbooks and programming languages, including mainstream ones like Java and Python (with nearly 2 million questions on Stack Overflow). While focused on programming education, our chain-of-illocution and evaluation methods can be adapted to other domains.

\paragraph{Construct Validity.} Our perceived correctness metric relies on accepted Stack Overflow answers, which may not always be optimal. Similarly, our adherence metrics, relying on LLM evaluations and the hyper-parameter \(t\), may not cover all aspects of the measured dimension and are subject to inaccuracies.  More broadly, we operationalize faithfulness as traceability to a single authoritative source (the textbook); this does not exhaust the space of explanation quality, which can also involve completeness, brevity, and pedagogical usefulness. In addition, the implicit-question set produced by chain-of-illocution is only an approximation of what a particular explainee needs; mismatches can lead to omitted context or unnecessary detail.

\section{Conclusion \& Future Work} \label{sec:conclusion}
We studied \ac{RAG}-based \ac{LLM} explanations through an XAI lens, focusing on \emph{source faithfulness}, i.e., whether the claims in an explanation are traceable to an explicit evidence source. Using 90 Stack Overflow questions spanning three programming textbooks, we find that current \ac{RAG} systems still exhibit only medium-to-low textbook adherence (\textbf{RQ1}), especially for larger models that tend to add plausible but ungrounded details. Grounded in Achinstein's illocutionary theory of explanation, we then introduce \textit{chain-of-illocution}: a prompting strategy that makes the explanation's macro-plan explicit via implicit guiding questions and shifts macro-planning control to retrieval. This improves source faithfulness (textbook adherence precision) by an average of 34\% across different \ac{LLM} families (\textbf{RQ2}), although the resulting adherence levels remain moderate and the gains are uneven across models.

Beyond these empirical gains, our main theoretical contribution is to operationalize Achinstein's illocutionary view of explanation as a practical design principle for explainable RAG: make the explanation macro-plan explicit (as implicit questions) and ground it in retrieval, while using the \ac{LLM} primarily for micro-planning and linguistic realization. A user study with 165 retained participants also found no evidence that chain-of-illocution reduces perceived explanation quality (\textbf{RQ3}).

Future work may expand chain-of-illocution to other educational contexts, integrate it with fine-tuning for enhanced adherence precision, and separate micro- and macro-planning via a mixture-of-experts pipeline \cite{xue2024openmoe,masoudnia2014mixture} to further strengthen adherence to authoritative textbooks.

\begin{credits}
\subsubsection{\ackname}
F. Sovrano and A. Bacchelli acknowledge the partial support of the Swiss National Science Foundation for the SNF Project 200021\_197227.

\end{credits} 

\bibliographystyle{splncs04nat}
\bibliography{references}

\end{document}